\documentclass[11pt,letterpaper]{article}
\usepackage{ijcnlp2017}
\usepackage{times}
\usepackage{latexsym}
\usepackage{qtree}
\usepackage{graphicx}
\usepackage{amsmath}
\usepackage{amsfonts}
\usepackage{multirow}
\usepackage{enumitem}
\ijcnlpfinalcopy

\def\ijcnlppaperid{321}

\newcommand{\figref}[1]{Figure \ref{#1}}
\newcommand{\tabref}[1]{Table \ref{#1}}
\newcommand{\secref}[1]{Section \ref{#1}}

\title{Tag-Enhanced Tree-Structured Neural Networks for Implicit Discourse Relation Classification}

\author{Yizhong Wang$^1$ \qquad Sujian Li$^{1,2}$ \qquad Jingfeng Yang$^1$ \qquad Xu Sun$^1$ \qquad Houfeng Wang$^{1,2}$\\
  	$^1$Key Laboratory of Computational Linguistics, Peking University, MOE, China \\
  	$^2$Collaborative Innovation Center for Language Ability, Xuzhou, Jiangsu, China\\  
  {\tt \{yizhong, lisujian, yjfllpyym, xusun, wanghf\}@pku.edu.cn} \\}

\date{}

\begin{document}

\maketitle

\begin{abstract}

Identifying implicit discourse relations between text spans is a challenging task because it requires understanding the meaning of the text. To tackle this task, recent studies have tried several deep learning methods but few of them exploited the syntactic information. In this work, we explore the idea of incorporating syntactic parse tree into neural networks. Specifically, we employ the Tree-LSTM model and Tree-GRU model, which are based on the tree structure, to encode the arguments in a relation. Moreover, we further leverage the constituent tags to control the semantic composition process in these tree-structured neural networks. Experimental results show that our method achieves state-of-the-art performance on PDTB corpus.
\end{abstract}

\section{Introduction}


It is widely agreed that text units such as clauses or sentences are usually not isolated. Instead, they correlate with each other to form coherent and meaningful discourse together. To analyze how text is organized, discourse parsing has gained much attention from both the linguistic \cite{weiss2007critical, tannen2012discourse} and computational \cite{soricut2003sentence,li2015tree,two-stage} communities, but the current performance is far from satisfactory. The most challenging part is to identify the discourse relations between text spans, especially when the discourse connectives (e.g., ``because'' and ``but'') are not explicitly shown in the text. Due to the absence of such evident linguistic clues, modeling and understanding the meaning of the text becomes the key point in identifying such implicit relations.

\begin{figure}[th]
\centering
\includegraphics[width=0.46\textwidth]{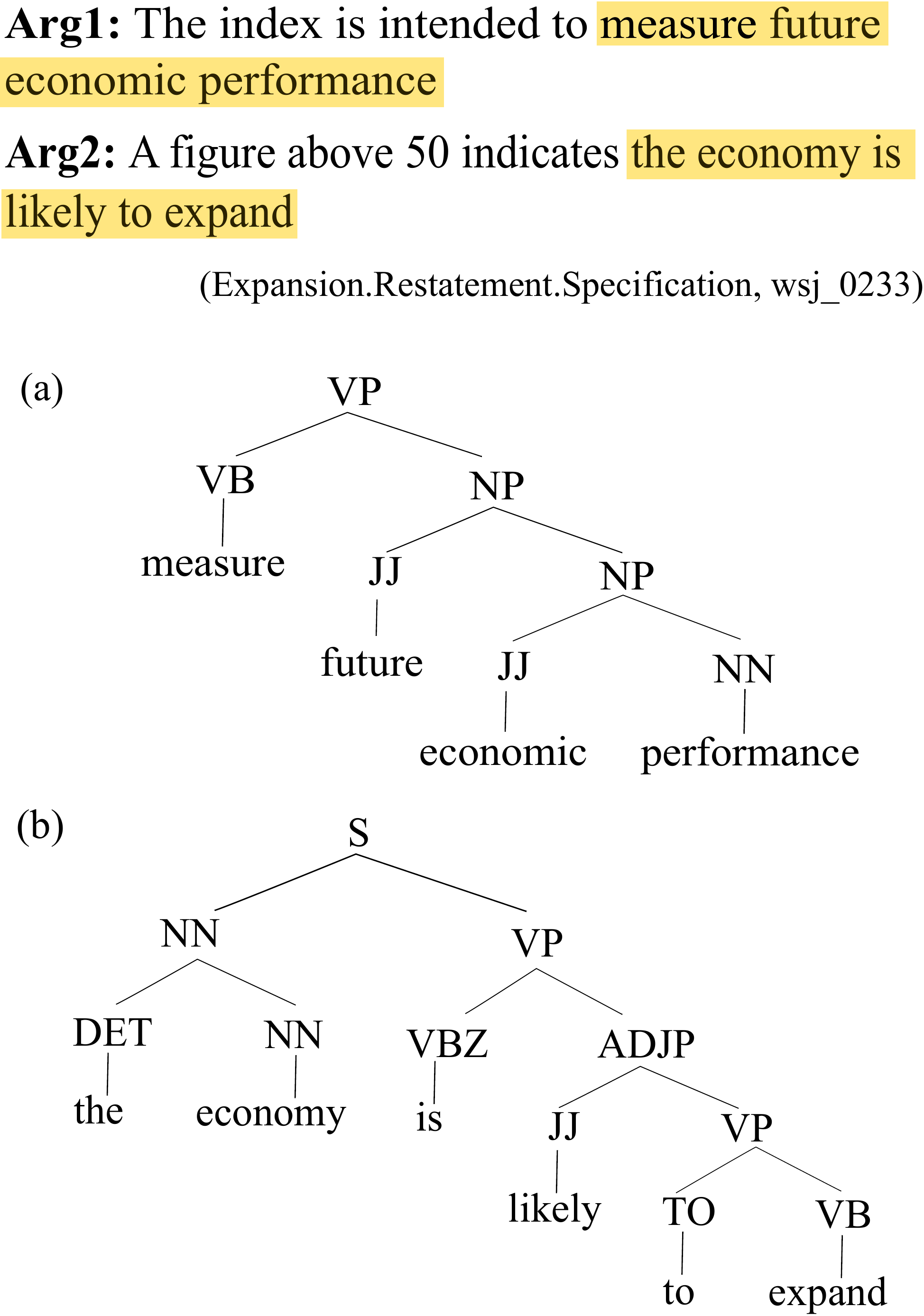}
\label{example}
\caption{An example of two sentences with their discourse relation as \textit{Expansion.Restatement.Specification}. Subfigure (a) and (b) are partial parse trees of the two important phrases with yellow background.}
\end{figure}

Previous studies in this field treat the task of recognizing implicit discourse relations as a classification problem and various techniques in semantic modeling have been adopted to encode the arguments in each relation, ranging from traditional feature-based models \cite{lin2009recognizing,pitler2009automatic} to the currently prevailing deep learning methods \cite{ji2014one, liu2016recognizing, qin2017adversarial}. Despite of the superior ability of the deep learning models, the syntactic information, which proves to be helpful for identifying discourse relations in many early studies \cite{subba2009effective, lin2009recognizing}, is seldom employed by recent work. Therefore we are curious to explore whether such missing syntactic information can be leveraged in deep learning methods to further improve the semantic modeling for implicit discourse relation classification.


Tree-structured neural networks, which recursively compose the representation of smaller text units into larger text spans along the syntactic parse tree, can tactfully combine syntatic tree structure with neural network models and recently achieve great success in several semantic modeling tasks \cite{eriguchi2016tree, kokkinos2017structural, chen2017enhanced}. One useful property of these models is that the representation of phrases can be naturally captured while computing the representations from bottom up. Taking \figref{example} for an example, those highlighted phrases could provide important signals for classifying the discourse relation. Therefore, we will employ two latest tree-structuerd models, i.e. the Tree-LSTM model \cite{tai2015improved, zhu2015long} and the Tree-GRU model \cite{kokkinos2017structural}, in our work. Hopefully, these models can learn to preserve or highlight such helpful phrasal information while encoding the arguments.


%


Another important syntactic signal comes from the constituent tags on the tree nodes (e.g., NP, VP, ADJP). Those tags, derived from the production rules, describe the generative process of text and therefore could indicate which part is more important in each constituent. For example, considering a node tagged with NP, its child node tagged with DT is usually neglectable. Thus we propose to incorporate this tag information into the tree-structured neural networks, where those constituent tags can be used to control the semantic compostion process.

Therefore, in this paper, we will approach the discourse relation classification task with two tree-structured neural networks proposed recently (Tree-LSTM  and Tree-GRU ). To our knowledge, this is the first time these models are applied to discourse relation classification. Moreover, we further enhance these models by leveraging the constituent tags to compute the gates in these models. Experiments on PDTB 2.0 \cite{DBLP:conf/lrec/PrasadDLMRJW08} show that the models we propose can achieve state-of-the-art results.

\section{Related Work}

\subsection{Implicit Discourse Relation Classiﬁcation}
\label{ss:implicit}

Discourse relation identification is an important but difficult sub-component of discourse analysis. One fundamental step forward recently is the release of the large-scale Penn Discourse TreeBank (PDTB) \cite{DBLP:conf/lrec/PrasadDLMRJW08}, which annotates discourse relations with their two textual arguments over the 1 million word Wall Street Journal corpus. The discourse relations in PDTB are broadly categorized as either ``Explicit'' or ``Implicit'' according to whether there are connectives in the original text that can indicate the sense of the relations. In the absence of explicit connectives, identifying the sense of the relations has proved to be much more difficult \cite{DBLP:conf/sigdial/ParkC12,DBLP:conf/eacl/RutherfordX14} since the inferring is solely based on the arguments. 

Prior work usually tackles this task of implicit discourse relation identification as a classification problem with the classes defined in PDTB corpus. Early attempts use traditional various feature-based methods and the work inspiring us most is \newcite{lin2009recognizing}, in which they show that the syntactic parse structure can provide useful signals for discourse relation classification. More specifically they employ the production rules with constituent tags (e.g., SBJ) as features and get competitive performance. Recently, with the popularity of deep learning methods, many cutting-edge models are also applied to our task of implicit discourse relation classification. \newcite{qin2016stacking} tries to model the sentences with Convolutional Neural Networks. \newcite{liu2016recognizing} encodes the text with Long Short Term Memory model and employ multi-level attention mechanism to capture important signals. \newcite{qin2017adversarial} proposes a framework based on adversarial network to incorporate the connective information. To be noted, \newcite{ji2014one} adopts Recursive Neural Network to exploit the representation of sentences and entities, which is the first yet simple tree-structured neural network applied in this task.

\subsection{Tree-Structured Neural Networks}

Tree-structured neural networks are one of the most widely-used deep learning models in natural language processing. Such neural networks usually recursively computes the representation of larger text spans from its constituent units according to the syntactic parse tree. Thanks to this compositional nature of text, tree-structured neural network models show superior ability in a variety of semantic modeling tasks, such as sentiment classification \cite{kokkinos2017structural}, natural language inference \cite{chen2017enhanced} and machine translation \cite{eriguchi2016tree}. 

The earliest and simplest tree-structure neural network is the Recursive Neural Network proposed by \newcite{socher2011parsing}, in which a global matrix is learned to linearly combine the contituent vectors. This work is further extended by replacing the global matrix with a global tensor to form the Recursive Neural Tensor Network \cite{socher2013recursive}. Based on them, \newcite{qian2015learning} first proposes to incorporate tag information, which is very similar as our idea described in \secref{ss:tag model}, by either choosing a composition function according to the tag of a phrase (Tag-Guided RNN/RNTN) or combining the tag embeddings with word embeddings (Tag-Embedded RNN/RNTN). Our method of incorporating tag information improves from theirs and somewhat combines these two methods by using the tag embedding to dynamically determine the composition function via the gates in LSTM or GRU.

One fatal weakness of vanilla RNN/RNTN is the well-known gradient exploding or vanishing problem due to the multiple computation steps in the vertical direction. Therefore \newcite{tai2015improved} and \newcite{zhu2015long} propose to import the Long Short Term Memory into tree structured neural networks and design a novel network architecture called Tree-LSTM. The adoption of the memory cell enables the Tree-LSTM model to preserve information even though the tree becomes very high. Similar to Tree-LSTM, \newcite{kokkinos2017structural} introduces the so-called Tree-GRU network, which replace the LSTM unit with Gated Recurrent Unit (GRU). With less parameters to train, Tree-GRU achieves better performance on the sentiment analysis task. In this work, we will experiment with these Tree-LSTM and Tree-GRU models for the semantic modeling in implicit relation classification.

\section{Our Method}
\label{method}

\begin{figure}[ht]
\centering
\includegraphics[width=0.46\textwidth]{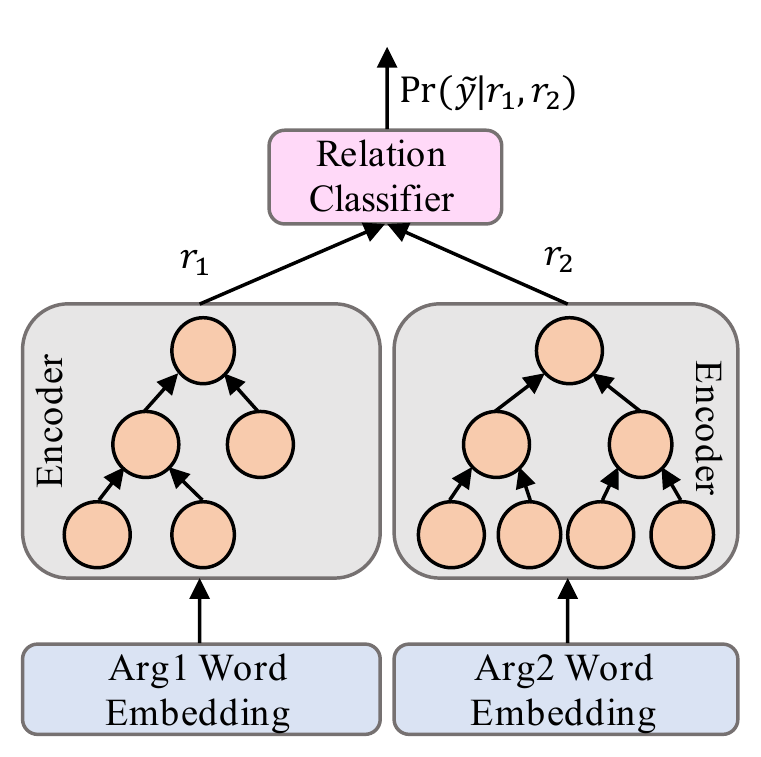}

\caption{Architecture of our discourse relation classification model. Layers with the same color share the same parameters.}
\label{fig:architecture}
\end{figure}

This section details the models we use for implicit discourse relation classification. Given two textual arguments without explicit connectives, our task is to classify the discourse relation between them. It can be viewed as two parts: 1) modeling the semantics of the two arguments; 2) classifying the relations based on the semantics. Our main contribution concentrates on the semantic modeling part with two types of tree-structured neural networks described in \secref{ss:tree model} and we further illustrate how to leverage the constituent tags to enhance these two models in \secref{ss:tag model}. In \secref{ss:classification}, we will shortly introduce the relation classifier and the training procedure of our model. The architecture of our system is illustrated in \figref{fig:architecture}.

\subsection{Modeling the Arguments with Tree-Structured Neural Networks}
\label{ss:tree model}

In a typical tree-structured neural network, given a parse tree of the text, the semantic representations of smaller text units are recursively composed to compute the representation of larger text spans and finally compute the representation for the whole text (e.g., sentence). In this work, we will construct our models based on the constituency parse tree, as is shown in \figref{example}. Following previous convention \cite{eriguchi2016tree,DBLP:journals/corr/ZhaoHM17}, we convert the general parse tree, where the branching factor may be arbitrary, into a binary tree so that we only need to consider the left and right children at each step. Then the following Tree-LSTM and Tree-GRU models can be used to obtain a vector representation of each argument.

\paragraph{Tree-LSTM Model.}
In a standard sequential LSTM model, the LSTM unit is repeated at each step to take the word at current step and previous output as its input, update its memory cell and output a new hidden vector. In the Tree-LSTM model, a similar LSTM unit is applied to each node in the tree in a bottom-up manner. Since each internal node in the binary parse tree has two children, the Tree-LSTM unit has to consider information from two preceding nodes, as opposed to the single preceding node in the sequential LSTM model. Each Tree-LSTM unit (indexed by $j$) contains an input gate $i_j$, a forget gate $f_j$ \footnote{The original Binary Tree-LSTM in \cite{tai2015improved} contains separate forget gates for different child nodes but we find single forget gate performs better in our task.} and an output gate $o_j$. The computation equations at node $j$ are as follows: 

\begin{align}
	i_j &= \sigma \left( W^{ \left( i \right)} x_j + U^{\left( i \right)} \left[ h_j^L, h_j^R \right] \right) \\
	f_j &= \sigma \left( W^{ \left( f \right)} x_j + U^{\left( f \right)} \left[ h_j^L, h_j^R \right] \right) \\
	o_j &= \sigma \left( W^{ \left( o \right)} x_j + U^{\left( o \right)} \left[ h_j^L, h_j^R \right] \right) \\
	u_j &= \tanh \left( W^{ \left( u \right)} x_j + U^{\left( u \right)} \left[ h_j^L, h_j^R \right]  \right) \\
	c_j &= i_j \odot u_j + f_j \odot c_j^L + f_j \odot c_j^R \\
	h_j &= o_j \odot \tanh \left(c_j\right) 
\end{align}

\noindent where $x_j$ is the embedded word input at current node $j$, $\sigma$ denotes the logistic sigmoid function and $\odot$ denotes element-wise multiplication. $h_j^L$, $h_j^R$ are the output hidden vectors of the left and right children, and $c_j^L$, $c_j^R$ are the memory cell states from them, respectively. To save space, we leave out all the bias terms in affine transformations and the same is true for other affine transformations in this paper.

Intuitively, $u_j$ can be regarded as a summary of the inputs at current node, which is then filtered by $i_j$. The memory from left and right children are forgotten by $f_j$ and then we compose them together with the new inputs to form the the new memory $c_j$. At last, part of the information in memory $c_j$ is exposed by $o_j$ to generate the output vector $h_j$ for current step. Another thing to note is that only leaf nodes in the constituency tree have words as its input, so $x_j$ is set to a  zero vector in other cases.

\paragraph{Tree-GRU Model.} 
Similar to Tree-LSTM, the Tree-GRU model extends the sequential GRU model to tree structures. The only difference between Tree-GRU and Tree-LSTM is how they modulate the flow of information inside the unit. Specifically, Tree-GRU unit removes the separate memory cell and only uses two gates to simulate the reset and update procedure in information gathering. The computation equations in each Tree-GRU unit are the following:

\begin{align}
	r_j &= \sigma \left( W^{\left( r \right)} x_j + U^{\left( r \right)} \left[ h_j^L, h_j^R \right] \right) \\
	z_j &= \sigma \left( W^{\left( z \right)} x_j + U^{\left( z \right)} \left[ h_j^L, h_j^R \right] \right) \\
	\tilde{h}_j &= \tanh \left( W^{\left( h \right)} x_j + U^{\left( h \right)} \left[ h_j^L \odot r_j, h_j^R \odot r_j \right] \right) \\
	h_j &= z_j \odot \tilde{h}_j + \left( 1 - z_j \right) \odot \left( h_j^{L} + h_j^{R} \right)
\end{align}

\noindent where $r_j$ is the reset gate and $z_j$ is the update gate. The reset gate allows the network to forget previous computed representations, while the update gate decides the degree of update to the hidden state. There is no memory cell in Tree-GRU, with only $h_j^L$ and $h_j^R$ as the hidden states from the left and right children.

\subsection{Controlling the Semantic Composition with Constituent Tags}
\label{ss:tag model}

The constituent tag in a parse tree describes the grammatical role of its corresponding constituent in the context. \newcite{bies1995bracketing} defines several types of constituent tags, including clause-level tags (e.g., SBAR, SINV, SQ), phrase-level tags (e.g., NP, VP, PP) and word-level tags (e.g., NN, VP, JJ). These constituent tags greatly interleave with the semantics and in some ways can provide determinant signals for the importance of a constituent. For example, for most of the time, constituents with PP (prepositional phrase) tag are less important than those with VP (verb phrase) tag. Therefore we argue that these tags are worth considering when we compose the semantics in the tree-structured neural networks.

One way to leverage such tags is using tag-specific composition functions but that would lead to large number of parameters and some tags are very sparse so it's very hard to train their corresponding parameters sufficiently.  To solve this problem, we propose to use tag embeddings and dynamically control the composition process via the gates in our model. 

Gates in Tree-LSTM and Tree-GRU units control the flow of information and thus determine how the semantics from child nodes are composed to a new representation. Furthermore, these gates are computed dynamically according to the inputs at a certain step. Therefore, it's natural to incorporate the tag embeddings in the computation of these gates. Based on this idea, we propose the Tag-Enhanced Tree-LSTM model, where the input, forget and output gates in each unit are calculated as follows: 

\begin{align}
	i_j &= \sigma \left( W^{ \left( i \right)} x_j + M^{\left( i \right)} t_j + U^{\left( i \right)} \left[ h_j^L, h_j^R \right]\right) \\
	f_j &= \sigma \left( W^{ \left( f \right)} x_j + M^{\left( f \right)} t_j + U^{\left( f \right)} \left[ h_j^L, h_j^R \right]\right) \\
	o_j &= \sigma \left( W^{ \left( o \right)} x_j + M^{\left( o \right)} t_j + U^{\left( o \right)} \left[ h_j^L, h_j^R \right]\right) 
\end{align}

Similarly, we can have the Tag-Enhanced Tree-GRU model with new reset and update gates:

\begin{align}
	r_j &= \sigma \left( W^{\left( r \right)} x_j + M^{\left( r \right)} t_j + U^{\left( r \right)} \left[ h_j^L, h_j^R \right]\right) \\
	z_j &= \sigma \left( W^{\left( z \right)} x_j + M^{\left( z \right)} t_j + U^{\left( z \right)} \left[ h_j^L, h_j^R \right] \right) 
\end{align}

\noindent where $t_j$ is the embedding of the tag at current  node (indexed by $j$).

\subsection{Relation Classification and Training}
\label{ss:classification}

In our work, the two arguments are encoded with the same network in order to reduce the number of parameters. After that we get a vector representation for each argument, which can be denoted as $r_1$ for argument 1 and $r_2$ for argument 2. Supposing that there are totally $n$ relation types, the predicted probability distribution $\hat{y}\in\mathbb{R}^n$ is calculated as:

\begin{equation}
	\hat{y} = softmax\left(W^{\left( \hat{y} \right)}\left[ r_1, r_2\right] + b^{\left( \hat{y} \right)}\right)
\end{equation} 

\begin{figure}[t]
\centering
\includegraphics[width=0.46\textwidth]{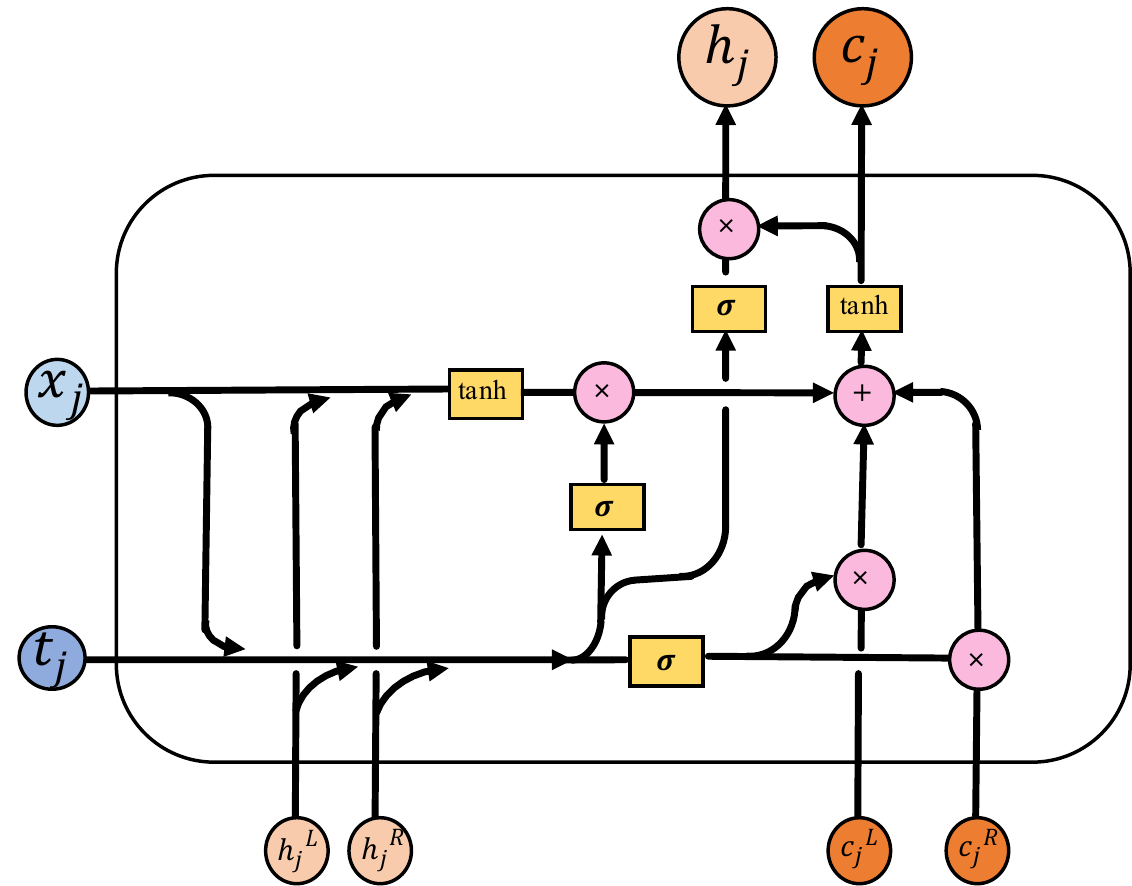}
\label{fig:tree-lstm}
\caption{Flow of Information in Tag-Enhanced Tree-LSTM unit. }
\end{figure}

\begin{figure}[t]
\centering
\includegraphics[width=0.46\textwidth]{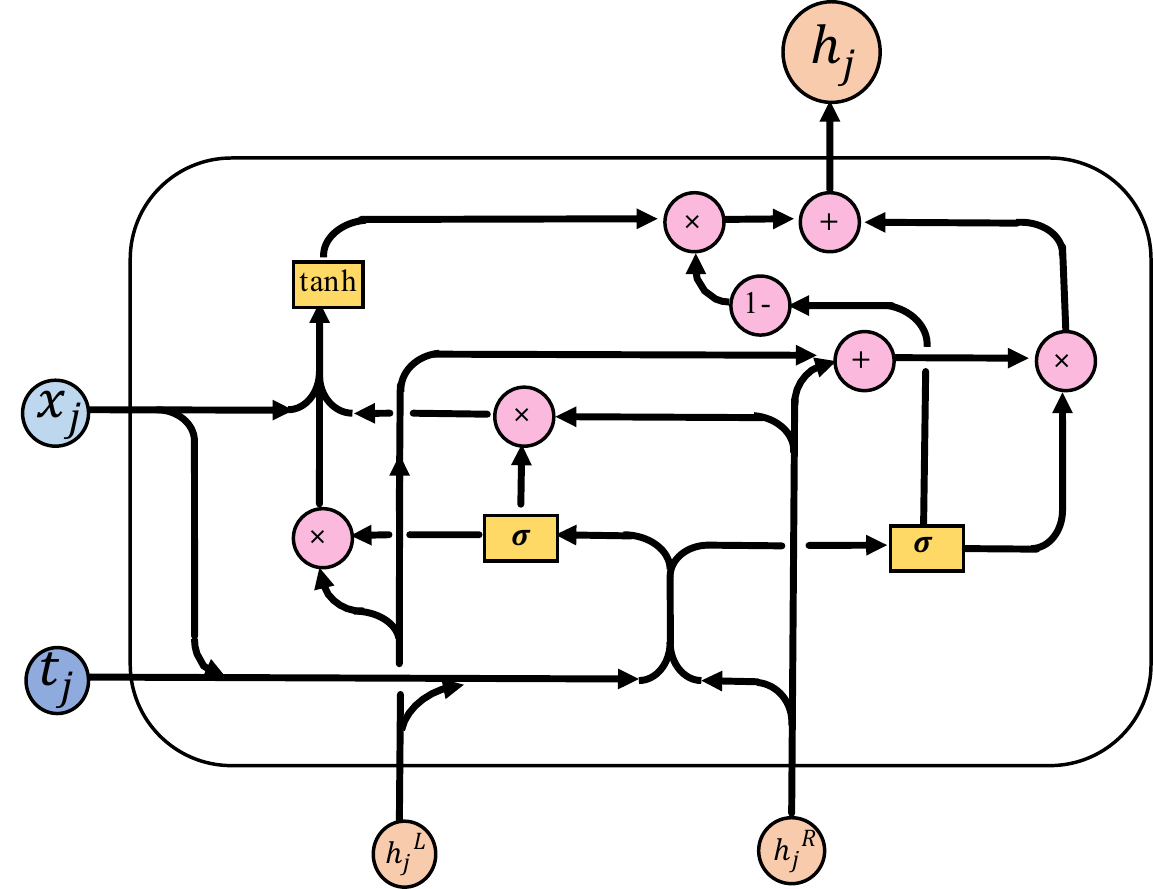}
\label{fig:tree-gru}
\caption{Flow of Information in Tag-Enhanced Tree-GRU unit. }
\end{figure}

To train our model, the training objective $J$ is deﬁned as the cross-entropy loss with $L2$ regularization:

\begin{align}
E\left(\hat{y}, y\right) &= - \sum_{j=1}^{n}y_j \times \log \hat{y}_j \\
J\left( \theta \right) &= \frac{1}{N} \sum_{k=1}^{N}	 E\left(\hat{y}, y\right) + \frac{\lambda}{2} {\|\theta\|}^2
\end{align}

\noindent where $\hat{y}$ is the predicted probability distribution, $y$ is the one-hot representation of the gold label and $N$ is the number of training samples.

\section{Experiments}

\subsection{Experiment Setup}

\begin{table*}[htbp]
\centering
\begin{tabular}{|c|c|c|c|c|}
\hline
\multirow{2}{*}{Models} & \multicolumn{2}{c|}{Level-1 Classification} & \multicolumn{2}{c|}{Level-2 Classification} \\ \cline{2-5} 
                        & \hspace{9pt} Dev \hspace{9pt}                 & Test                  & \hspace{9pt} Dev \hspace{9pt}                 & Test                  \\ \hline
Bi-LSTM                 &    55.10                 & 56.88                 &    35.02                 & 42.44                 \\ \hline

Bi-GRU                  &    55.21                 & 57.01                 &    35.34                 & 42.46                 \\ \hline \hline
Tree-LSTM               &     56.04                & 58.89                 & 35.76                    & 43.02                 \\ \hline
Tree-GRU                &      55.36               & 58.98                 &   36.09                  & 43.78                 \\ \hline \hline
Tag-Enhanced Tree-LSTM  &       \textbf{56.97}              & \textbf{59.85}                 &   35.92                 & \textbf{45.21}                 \\ \hline
Tag-Enhanced Tree-GRU   &       56.63              & 59.75                 &   \textbf{36.93}                  & 44.55                 \\ \hline
\end{tabular}
\caption{The accuracy score of multi-class classification}\label{tab:perf}
\end{table*}

\paragraph{Corpus.}
We evaluate our method on the Penn Discourse Treebank (PDTB) \cite{DBLP:conf/lrec/PrasadDLMRJW08}, which provides annotations of discourse relations over the Wall Street Journal corpus. Each relation instance consists of two arguments, typically adjacent pairs of sentences in a text. As is mentioned in \secref{ss:implicit}, the relations in PDTB are generally categorized into either explicit or implicit and our work focuses on the more challenging implicit relation classification task. Totally, there are 16,224 implicit relation instances in PDTB dataset, with a three-level hierarchy. The first level is defined as 4 major classes of the relation, including: \textit{Temporal}, \textit{Contingency}, \textit{Comparison} and \textit{Expansion}. Then for each class, it is further divided into different types, which is supposed to provide finer pragmatic distinctions. This totally yields 16 relation types at the second level. At last, a third level of subtypes is defined for some types according to the semantic contribution of each argument. 

\paragraph{Preprocesssing.}
Following the common setup convention \cite{DBLP:conf/eacl/RutherfordX14, ji2014one, liu2016recognizing}, we split the dataset into training set (Sections 2-20), development set (Sections 0-1), and test set (Section 21-22). For preprocessing, we employ the Stanford CoreNLP toolkit \cite{manning-EtAl:2014:P14-5} to lemmatize all the words and get the constituency parse tree for each sentence. Then we convert the parse tree into binary with right branching and we remove the internal nodes that have only one child so that our binary tree-structured models can be applied. Small portion of the arguments in PDTB are composed of multiple sentences. In such cases, we add a new ``Root'' node and link the original root nodes of those sentences to this shared ``Root'' before converting the tree into binary. 

\paragraph{Multi-Class Classification}There are mainly two ways to set up the classification tasks in previous work. Early studies \cite{pitler2009automatic, DBLP:conf/sigdial/ParkC12} train and evaluate separate ``one-versus-all'' classifiers for each discourse relation since the classes are extremely imbalanced in PDTB. However, recent work put more emphasis on the multi-class classification, where the goal is to identify a discourse relation from all possible choices. According to \newcite{DBLP:conf/eacl/RutherfordX14}, the multi-class classification setting is more natural and realistic. Moreover, the multi-class classifier can directly serve as one building block of a complete discourse parser \cite{qin2017adversarial}. Therefore, in this work, we will focus on the multi-class classification task. Moreover, we  will experiment on the classification of both the Level-1 classes and the Level-2 types, so that we can compare with most of previous systems and thoroughly analyze the performance of our method.

It should be noted that roughly 2\% of the implicit relation instances are annotated with more than one types. Following \newcite{ji2014one}, we treat these different types as multiple instances while training our model. During testing, if the classifier hits either of the annotated types, we consider it to be correct. We also follow previous studies \cite{lin2009recognizing, ji2014one} to remove the instances of 5 very rare relation types in the second level. Therefore we have totally 11 types to classify for Level-2 classification and 4 classes for Level-1 classification.

\subsection{Model Settings}

We tune the hyper-parameters of our Tag-Enhanced Tree-LSTM model based on the development set and other models share the same set of hyper-parameters. The best-validated hyper-parameters, including the size of the word embeddings $\omega$, the size of the tag embeddings $\tau$, the dimension of the Tree-LSTM or Tree-GRU hidden state $d$, the learning rate $\eta$, the weight of L2 regularization term $\lambda$ and the batch size $b$ are shown in \tabref{tab:hyper}.

\begin{table}[htbp]
\centering
\begin{tabular}{|c|c|c|c|c|c|}
\hline 
$\omega$ & $\tau$ & $d$ & $\eta$ & $\lambda$ & $b$ \\ 
\hline 
50 & 50 & 250 & 0.01 & 0.0001 & 10 \\ 
\hline 
\end{tabular} 
\caption{Hyper-parameters of our model}\label{tab:hyper}
\end{table}

The Pre-trained 50-dimentional Glove Vectors \cite{pennington2014glove}, which is case-insensitive, are used for initializing the word embeddings and they are tuned together with other parameters in the same learning rate during training. 

We adopt the AdaGrad optimizer \cite{duchi2011adaptive} for training our model and we validate the performance every epoch. It takes around 5 hours (5 epochs) for the Tag-Enhanced Tree-LSTM and 4 hours (6 epochs) for the Tag-Enhanced Tree-GRU model to converge to the best performance, using one INTEL(R) Core(TM) I7 3.4GHz CPU and one NVIDIA GeForce GTX 1080 GPU.

\subsection{Results}

The evaluation results of our models on both the Level-1 classification and the Level-2 classification are reported in \tabref{tab:perf}. Accuracy score is used to measure the overall performance and we present our performance on both the development set and the test set. In addition to the four tree-structured neural networks  described in \secref{method}, we also implement two baseline models: the bi-directional LSTM model and the bi-directional GRU model. The hyper-parameters of these two models are tuned separately from other models. Due to the space limitation, we don't present the details here. Comparison of these sequential models with the tree-structured models are expected to show the effects of tree structures.

From \tabref{tab:perf}, we can see that the sequential Bi-LSTM model and Bi-GRU model perform worst for our task, which confirms our hypothesis that the tree-structured neural networks can really capture some important signals that are missing in the sequential models. 

Furthermore, if we add the tag information to tree-structured models, both the Tag-Enhanced Tree-LSTM model and the Tag-Enhanced Tree-GRU model provide conspicuous improvement (around 1\%) compared with the no-tag version. This demonstrates the usefulness of those constituent tags and the effectiveness of our method to incorporate this important feature.  Especially, since both the Tree-LSTM model and Tree-GRU model rely on the gating machanism to control the flow of information, this double confirms that the tag information can help with the computation of such gates and therefore can be leveraged to control the semantic composition process. 

Another discovery from our results is that the GRU models performs similarly as its corresponding variant of LSTM model. This conforms to previous empirical observation in sequential models that LSTM and GRU have comparable capability \cite{DBLP:journals/corr/ChungGCB14}. However, the Tree-GRU model have less parameters to train which could alleviate the problem of overfitting and also cost less training time.

\subsection{Comparison with Other Systems}

\begin{table}[t!]
\centering
\begin{tabular}{c|c}

\hline
Systems & Accuracy \\ \hline
\newcite{DBLP:conf/emnlp/ZhangSXLDY15} & 55.39  \\ 
\newcite{DBLP:conf/eacl/RutherfordX14}        & 55.50      \\ 
\newcite{DBLP:conf/naacl/RutherfordX15}     & 57.10      \\
\newcite{liu2016implicit}   & 57.27     \\
\newcite{liu2016recognizing} & 57.57 \\
\newcite{ji2016latent} & 59.50 \\
 \hline
Tag-Enhanced Tree-LSTM  & \textbf{59.85} \\
Tag-Enhanced Tree-GRU  & 59.75 \\ \hline
\end{tabular}
\caption{Accuracy (\%) for Level-1 multi-class classification on the test set, compared with other state-of-the-art systems.}
\label{tab:level1}
\end{table}

\begin{table}[t!]
\centering
\begin{tabular}{c|c}
\hline
Systems & Accuracy \\ \hline
\newcite{lin2009recognizing} & 40.66  \\ 
\newcite{ji2014one}        & 44.59      \\ 
\newcite{qin2016stacking}     & 45.04      \\
\newcite{qin2017adversarial}   & \textbf{46.23}     \\ \hline
Tag-Enhanced Tree-LSTM  & 45.21 \\
Tag-Enhanced Tree-GRU  & 44.55 \\ \hline
\end{tabular}
\caption{Accuracy (\%) for Level-2 multi-class classification on the test set, compared with other state-of-the-art systems.}
\label{tab:level2}
\end{table}

For a comprehensive study, we compare our models with other state-of-the-art systems.  The systems that conduct Level-1 classification are reported in \tabref{tab:level1}, including:

\begin{itemize}
	\item \newcite{DBLP:conf/emnlp/ZhangSXLDY15} proposes to use convolutional neural networks to encode the arguments.
	\item \newcite{DBLP:conf/eacl/RutherfordX14} manually extracts features to represent the arguments and use a maximum entropy classifier for classification. \newcite{DBLP:conf/naacl/RutherfordX15} further exploits discourse connectives to enrich the training data.
	\item \newcite{liu2016implicit} employs a multi-task framework that can leverage other discourse-related data to help with the training of discourse relation classifier.
	\item \newcite{liu2016recognizing} represents arguments with LSTM and introduces a multi-level attention mechanism to model the interaction between the two arguments.
	\item \newcite{ji2016latent} treats the discourse relation as latent variable and proposes to model them jointly with the sequences of words using a latent variable recurrent neural network architecture.\end{itemize}

And in \tabref{tab:level2}, we present the following systems, which focus on Level-2 classification:

\begin{itemize}
	\item \newcite{lin2009recognizing} uses traditional feature-based model to classify relations. Especially, constituent and dependency parse trees are exploited.
	\item \newcite{ji2014one} models both the semantics of argument and the meaning of entity mention with two recursive neural networks, which are then combined to classify relations.
	\item \newcite{qin2016stacking} utilize convolutional neural network for argument modeling and a collaborative gated neural network to model their interaction.
	\item \newcite{qin2017adversarial} proposes to incorporate the connective information via a novel pipelined adversarial framework. 
\end{itemize}

The comparison with these latest work shows that our system achieves currently best performance for the Level-1 classification and ranks second for the Level-2. With the state-of-the-art performance on both levels, we can verify the effectiveness of our method.

\subsection{Qualitative Analysis}
\label{analysis}

\begin{figure}[bht]
\centering
 \includegraphics[width=0.48\textwidth]{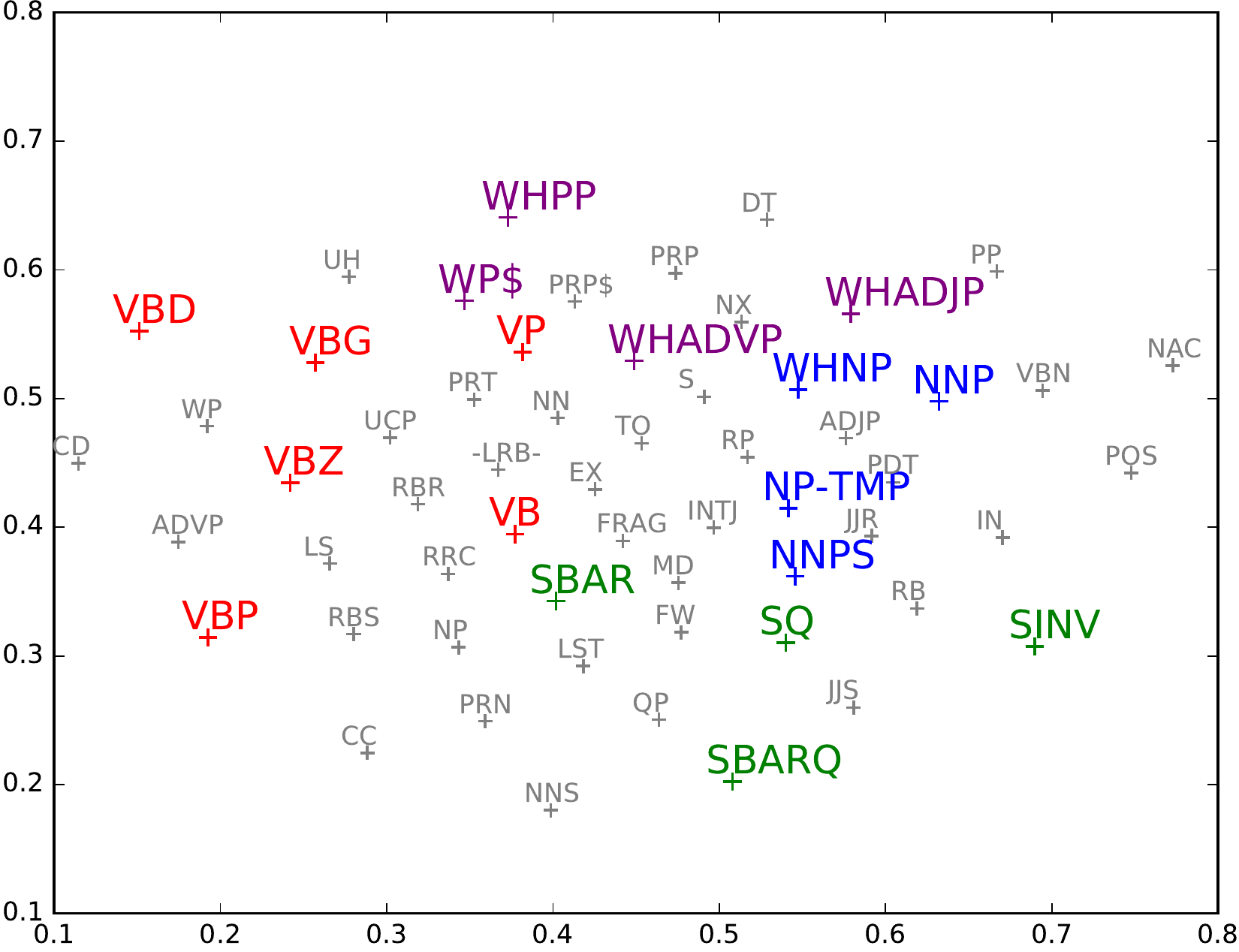}

\caption{t-SNE Visualization \cite{maaten2008visualizing} of the constituent tag embeddings}
\label{fig:tsne}
\end{figure}

To get a deeper insight into the proposed tag-enhanced models, we project the constituent tag embeddings learned in our Tag-Enhanced Tree-LSTM model into two dimensions using the t-SNE method \cite{maaten2008visualizing} and normalize the values in each dimension. The projected embeddings are visualized in \figref{fig:tsne} and we highlight some representative tags that may share some kind of commonality in their functions.

According to the definition in \newcite{bies1995bracketing}, the tags with red color are all verb-related, while those with blue color describes noun-related syntax. Despite of some noisy points, we can see that they are roughly separated into two groups. In addition, the purple tags correspond to words or phrase that are wh-related (e.g., what, when, where, which) and we can see they are distributed similarly. Moreover, the green tags are, which are located in the right-bottom corner, all describe the clause-level constituents. 

Therefore, from this visualization, we can conclude that the tag embeddings we learned are somewhat meaningful and really capture some functionalities of these tags. Since these embeddings are used to compute the gates and the gates further determine the flow of information, we argue that these tags can indeed help to control the semantic composition process in our tree-structured networks.

\section{Conclusion}

In this work, we propose to use two latest tree-structured neural networks to model the arguments for discourse relation classification. The syntactic parse tree are exploited from two aspects: first, we leverage the tree structure to recursively compose semantics in a bottom-up manner; second, the constituent tags are used to control the semantic composition process at each step via gating mechanism. Comprehensive experiments show the effectiveness of our proposed method and our system achieves state-of-the-art performance for the challenging task of implicit discourse relation classification. For future work, we will try other types of syntax embeddings and we are also working on incorporating structural attention mechanism into our tree-based models.

\section*{Acknowledgments}
We thank all the anonymous reviewers for their insightful comments on this paper.
This work was partially supported by National Natural Science Foundation of China (61572049 and 61333018).
The correspondence author of this paper is Sujian Li.

%

\bibliography{ijcnlp2017}
\bibliographystyle{ijcnlp2017}

\end{document}